\documentclass[10pt,leqno,twocolumn]{amsart}
\usepackage{graphicx}
\usepackage[margin=1in]{geometry}
\usepackage{changepage}
\usepackage{microtype}
\usepackage{indentfirst,csquotes}
\usepackage{float}
\usepackage{amssymb,amsthm,amsmath}
\usepackage{xcolor,paralist,hyperref,titlesec,fancyhdr,etoolbox}
\usepackage{booktabs}
\usepackage{multirow}

\titleformat{\section}
  {\normalfont\Large\bfseries}
  {\thesection.}{1em}{}

\titleformat{\subsection}
  {\normalfont\large\bfseries}
  {\thesubsection.}{1em}{}
  
\titleformat{\subsubsection}
  {\normalfont\normalsize\bfseries}
  {\thesubsubsection.}{1em}{}

\titlespacing*{\section}{0pt}{2ex}{1ex}
\titlespacing*{\subsection}{0pt}{1.5ex}{0.5ex}

\hypersetup{ colorlinks=true, linkcolor=black, filecolor=black, urlcolor=blue, citecolor=black }

\begin{document}
\twocolumn[
\sloppy
\begin{center}
    University of Tartu
    \vspace{6pt}
    
    {\Large \textbf{Can LLMs Assess Personality?\\ Validating Conversational AI for Trait Profiling}} \\
    \vspace{12pt}
    
    Andrius Mat\v{s}enas, Anet Lello, T\~{o}nis Lees, Hans Peep, Kim Lilii Tamm

\end{center}

\vspace{10pt}

\begin{adjustwidth}{3cm}{3cm}
\small
{\centering \textbf{Abstract} \par}
\vspace{0.5em}
    This study validates Large Language Models (LLMs) as a dynamic alternative to questionnaire-based personality assessment. Using a within-subjects experiment (N=33), we compared Big Five personality scores derived from guided LLM conversations against the gold-standard IPIP-50 questionnaire, while also measuring user-perceived accuracy. Results indicate moderate convergent validity (\textit{r}=0.38–0.58), with Conscientiousness, Openness, and Neuroticism scores statistically equivalent between methods. Agreeableness and Extraversion showed significant differences, suggesting trait-specific calibration is needed. Notably, participants rated LLM-generated profiles as equally accurate as traditional questionnaire results. These findings suggest conversational AI offers a promising new approach to traditional psychometrics.
\end{adjustwidth}
\vspace{10pt}
]

\section{Introduction}

Personality assessments are used in consequential decision-making processes such as hiring, dating, therapy and education. Yet these assessments rely heavily on explicit self-reporting through static questionnaires. While these methods are established, they often fail to capture the nuance, fluidity, and contextual nature of human behavior, limiting the depth of psychological profiling \cite{dyrenforth2010}. 

Large Language Models offer a potential solution. Their conversational capabilities enable dynamic, adaptive information gathering that can probe deeper when responses are ambiguous and adjust to individual communication styles. Instead of rigid self-evaluation, LLM-based assessment could extract personality signals from natural dialogue.

Recent work demonstrates that LLMs can infer Big Five traits from existing text corpora such as social media posts \cite{peters2024} and interview transcripts \cite{dai2022} \cite{park2024}. However, these approaches analyze static data rather than conducting real-time assessment.

This paper addresses this gap. We validate whether LLMs can extract Big Five personality traits from guided conversation with accuracy comparable to standardized questionnaires, such as IPIP-50 \cite{ipip}. We focus on the following research questions:

\begin{enumerate}
    \item Do LLM-extracted Big Five personality trait scores correlate with results from standardized IPIP-50 questionnaires?
    \item Do users perceive LLM-generated personality profiles extracted from casual conversation as accurate representations of themselves?
\end{enumerate}

We aim to contribute a validation framework for conversational psychological profiling suitable for consumer applications such as hiring, relationships, or personalized recommendation systems.

\section{Related Work}

\subsection{Traditional Personality Assessment}

The Big Five model (Openness, Conscientiousness, Extraversion, Agreeableness, Neuroticism) \cite{mccrae1992} represents the dominant framework in personality psychology. The IPIP-50 questionnaire \cite{ipip} provides a validated, public-domain instrument for measuring these traits through 50 self-report items (10 per trait). However, such questionnaires suffer from known limitations including social desirability bias, response fatigue, and inability to capture contextual behavioral variation.

\subsection{LLM-Based Psychological Profiling}

Recent research demonstrates that LLMs can infer personality from text with varying degrees of accuracy. Peters \& Matz \cite{peters2024} showed that GPT-3.5 and GPT-4 can infer Big Five traits from Facebook posts in a zero-shot manner, achieving correlations of $r = 0.29-0.33$ with self-reported scores. GPT-4 outperformed GPT-3.5, with Openness and Extraversion being the easiest traits to predict.

Piastra \& Catellani \cite{piastra2025} demonstrated that Big Five traits can be inferred from written text with ChatGPT-4 resulting in small-to-moderate correlations ($r \approx 0.25-0.40$). However, confidence scores provided by the model were poorly aligned with actual accuracy, indicating limitations in autonomous psychological assessment without additional validation.

Liu et al. \cite{liu2025} reported that LLM predictions showed strong linear alignment with human data ($R^2 > 0.89$) on inter-scale correlations, substantially exceeding predictions based on semantic similarity and approaching the accuracy of supervised machine learning algorithms.

\subsection{Conversational Personality Inference}

Moving beyond static text analysis, Dai, Jayaratne, \& Jayatilleke \cite{dai2022} developed InterviewBERT for predicting HEXACO personality traits from over 3 million job interview transcripts. They achieved an average correlation of $r = 0.37$ (with Openness reaching $r = 0.45$), demonstrating that conversational data can yield valid personality inferences.

Park et al. \cite{park2024} created LLM agents from two-hour interviews with 1,052 real people. These agents achieved 85\% accuracy replicating participant responses and $r = 0.95$ correlation on Big Five scores that matches human two-week test-retest reliability. This noteworthy finding suggests that extended conversational interaction can capture personality with high fidelity.

Ramon et al. \cite{ramon2021} demonstrated that behavioral patterns captured by machine learning models correlate with interpretable human attributes, supporting the theoretical basis that conversational behavior reflects systematic, measurable patterns of cognition and personality.

\subsection{Research Gap}

While prior work validates LLM personality inference from existing text corpora or extended interviews, there remains limited validation of real-time conversational assessment against gold-standard psychometric instruments, particularly with explicit measures of user-perceived accuracy. Our study addresses this gap by comparing guided LLM conversations directly with IPIP-50 scores while also capturing participants' subjective accuracy ratings.

\section{Methods}

\subsection{Study Design}

\begin{figure}[htbp]
    \centering
    \includegraphics[width=\columnwidth]{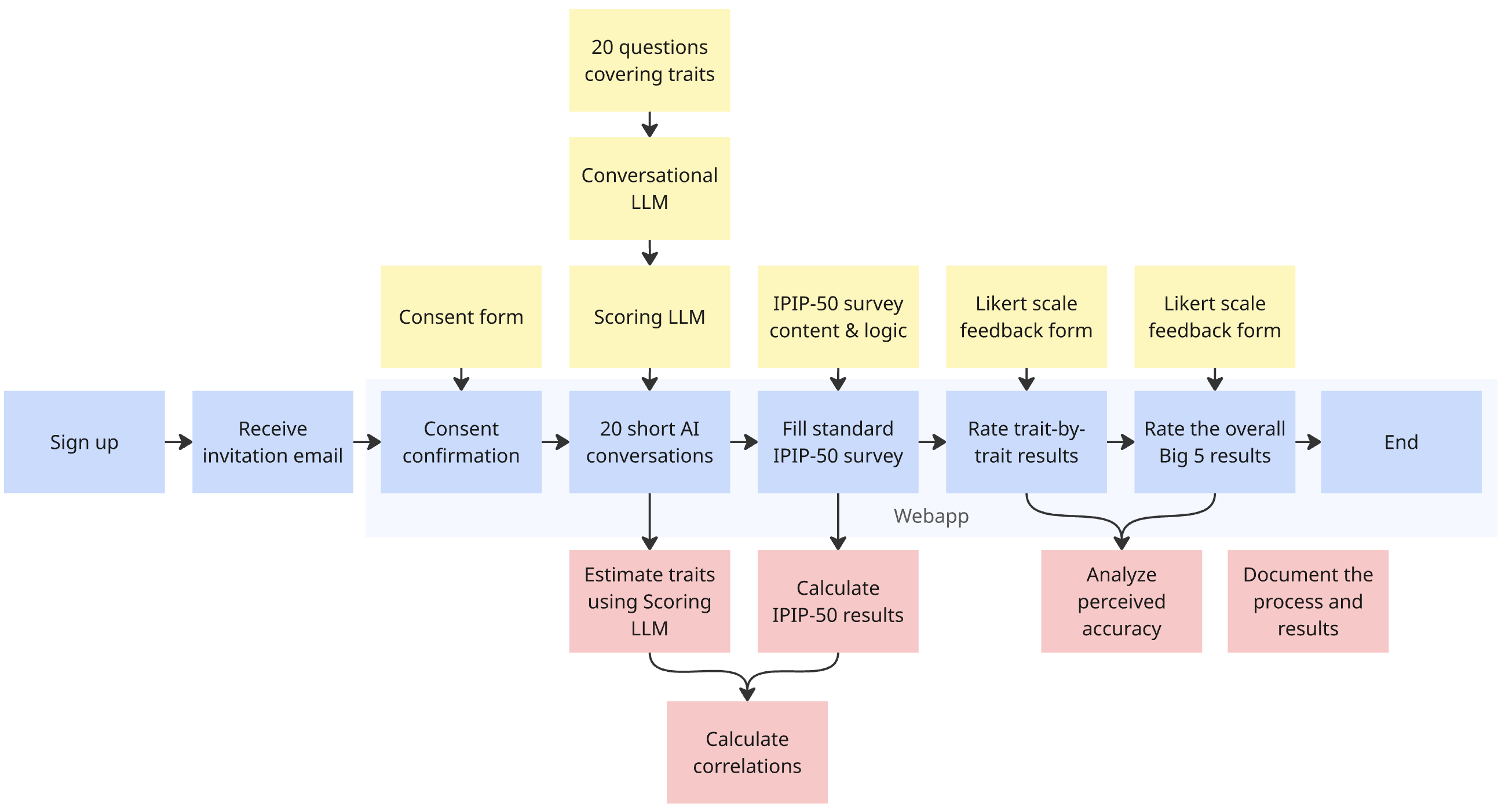}
    \caption{Experiment design}
    \label{fig:design_graph}
\end{figure}

We employed a within-subjects study design comparing two personality assessment methods: (1) constrained conversations with a general-use LLM (Gemini 2.5 Flash) and (2) the standard IPIP-50 questionnaire. This design, as seen in Figure~\ref{fig:design_graph}, allows direct comparison of scores from both methods for each participant, controlling for individual differences.

\subsection{Participants}

We aimed to recruit 100 participants, with an absolute minimum target of 30 to ensure adequate statistical power for detecting medium-to-large effects ($d \approx 0.5-0.6$) in paired comparisons. Participants were recruited through social media and university channels. Final data collection yielded 33 complete participations.

Basic demographic information and prior experience with chatbots and personality tests were collected during registration. Every participant reviewed an ethics statement explaining the study's purpose, participants' rights, and data handling procedures at the start of the experiment. Further details on the ethics statement can be found in Appendix B.

\subsection{Materials}

\subsubsection{IPIP-50 Questionnaire}

The International Personality Item Pool 50-item inventory \cite{ipip} served as our gold-standard measure. It contains 10 items per Big Five domain, with a mix of positively and negatively keyed items. Participants rate statements on a 5-point Likert scale from ``Very Inaccurate'' to ``Very Accurate.''

\subsubsection{Conversational Question Battery}

We designed 20 open-ended questions (4 per Big Five trait) to elicit personality-relevant responses through natural conversation. Questions were crafted to invoke behavioral descriptions and preferences without explicitly mentioning personality constructs. Example questions include: \\
\textbf{Openness:} ``Where would you like to travel to? Why that place?'' \\
\textbf{Conscientiousness:} ``What’s one important thing you had to get done this week? How did it go?'' \\
\textbf{Extraversion:} ``After a long, exhausting day, would you rather spend time alone or call a friend? Why?'' \\
\textbf{Agreeableness:} ``Tell me about the last time someone asked you for help or a favor'' \\
\textbf{Neuroticism:} ``Tell me about a recent mistake you made. How did you handle it?''

Each question was paired with LLM guidance specifying exploration depth and termination criteria.

\subsubsection{LLM Configuration}

A conversational LLM received a unique system prompt for each question, guiding it to explore relevant behavioral indicators. Conversations ended dynamically when the LLM determined sufficient information had been provided, or after a maximum of 10 user messages to prevent participant fatigue.

A separate scoring LLM analyzed all 20 conversation transcripts together and estimated personality trait scores on a 0-120 scale, which we normalized to 0-100 for visualization and interpretability.

Appendix A outlines the entire system prompt architecture including further details on question design and the implementation of variable prompts.

\subsection{Procedure}

\begin{figure}[htbp]
    \includegraphics[width=\columnwidth]{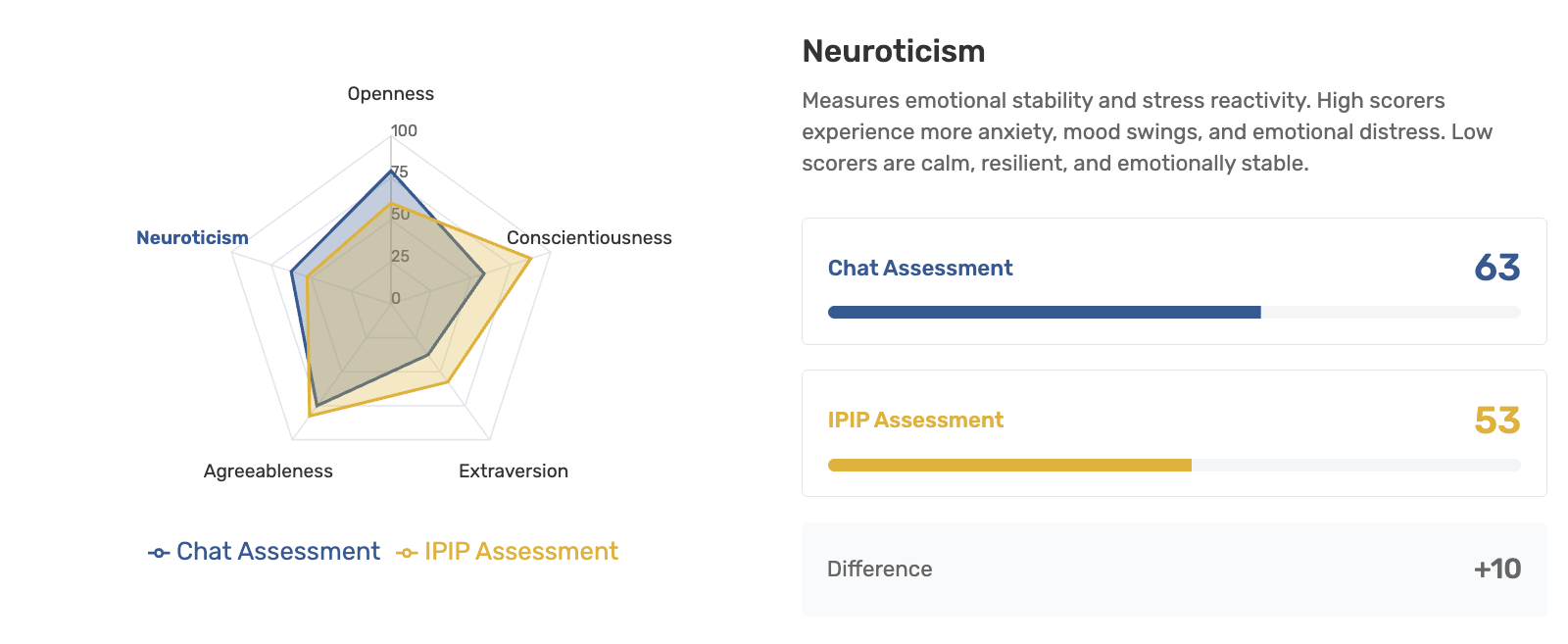}
    \caption{Example spider graph showing Big Five personality trait scores from both methods, displayed to participants before final feedback.}
    \label{fig:spider}
\end{figure}

Participants completed the study through a custom web application. The experiment followed these steps:

\begin{enumerate}
    \item Informed consent
    \item 20 brief LLM conversations
    \item IPIP-50 questionnaire completion
    \item Rating result scores (each trait individually to avoid bias)
    \item Viewing full results with trait explanations (See Figure~\ref{fig:spider})
    \item Rating overall method preference
\end{enumerate}

\subsection{Analysis Plan}

Given the within-subjects approach, we employed paired-samples $t$-tests to compare IPIP and LLM scores for each trait. We computed Pearson correlations to assess convergent validity between methods. We used Principal Component Analysis (PCA) to explore the underlying structure across all measures. Hierarchical clustering examined whether same-trait scores from different methods clustered together. For perceived accuracy, we compared Likert ratings between methods using paired $t$-tests.

\section{Results}

\subsection{Score Comparisons}

Table~\ref{tab:ttests} presents the results of paired $t$-tests comparing trait scores between methods. Agreeableness ($p = 0.002$) and Extraversion ($p = 0.036$) showed statistically significant differences between methods, indicating that IPIP and Chat produce meaningfully different estimates for these traits. For Conscientiousness ($p = 0.084$), Openness ($p = 0.485$), and Neuroticism ($p = 0.441$), no significant differences were observed, suggesting strong consistency between the two measurement approaches.

\begin{table}[H]
\centering
\begin{tabular}{lcc}
\toprule
\textbf{Trait} & \textbf{$p$ (scores)} & \textbf{$p$ (accuracy)} \\
\midrule
Agreeableness & 0.002* & 0.745 \\
Extraversion & 0.036* & 0.144 \\
Conscientiousness & 0.084 & 0.297 \\
Openness & 0.485 & 0.586 \\
Neuroticism & 0.441 & 0.686 \\
\bottomrule
\multicolumn{3}{l}{\small * Significant at $\alpha = 0.05$}
\end{tabular}
\caption{Paired $t$-test results: $p$-values for trait scores and perceived accuracy ratings}
\label{tab:ttests}
\end{table}

\begin{figure}[htbp]
    \includegraphics[width=\columnwidth]{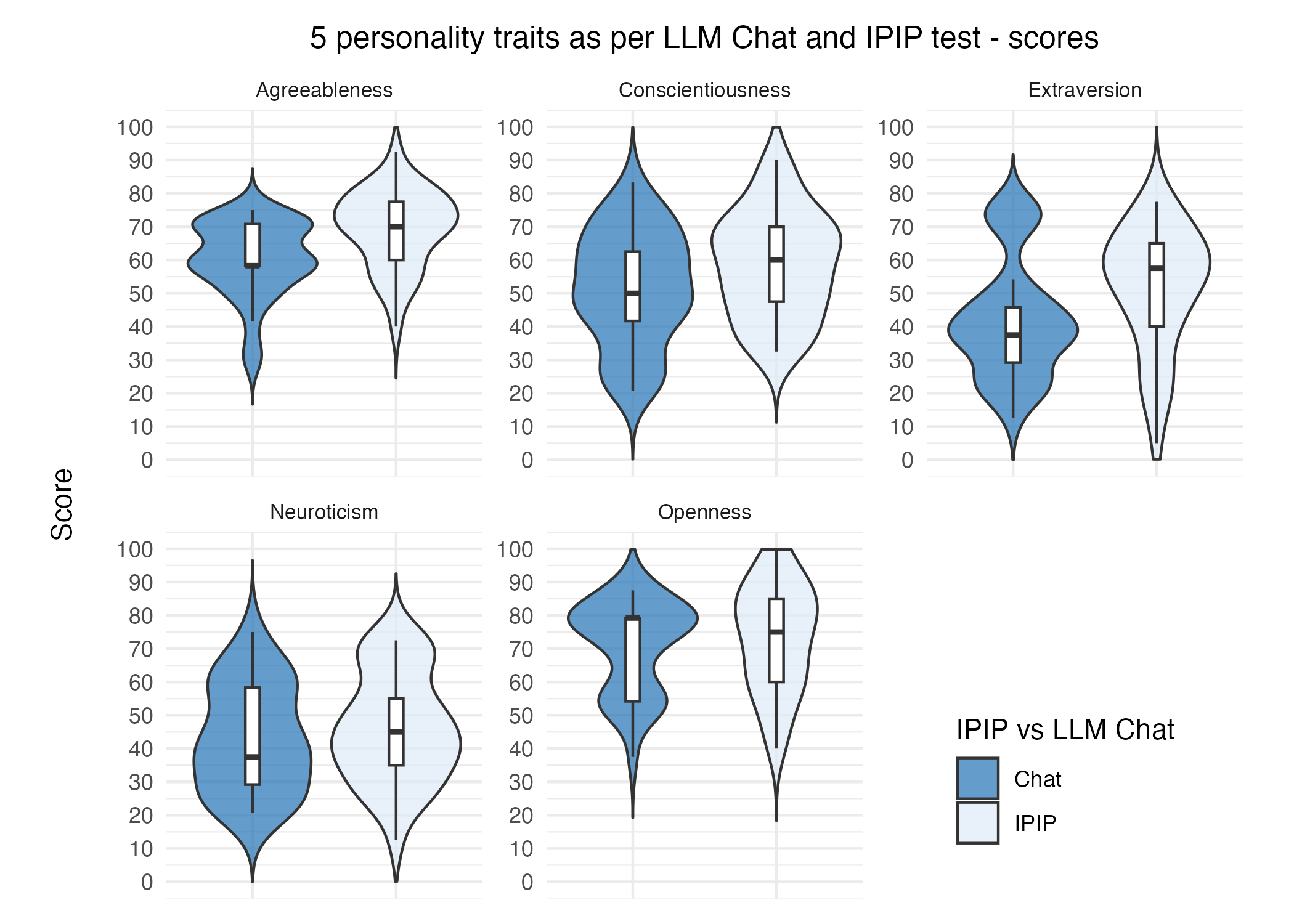}
    \caption{Distribution of Big Five trait scores by assessment method.}
    \label{fig:scores}
\end{figure}

Score distributions (Figure~\ref{fig:scores}) revealed that Openness and Extraversion showed wider variance in Chat results, while IPIP scores were more uniformly distributed.

\subsection{Convergent Validity}

Table~\ref{tab:correlations} presents correlations between Chat and IPIP scores for each trait.

\begin{table}[H]
\centering
\begin{tabular}{lc}
\toprule
\textbf{Trait} & \textbf{$r$} \\
\midrule
Extraversion & 0.58 \\
Openness & 0.46 \\
Neuroticism & 0.44 \\
Conscientiousness & 0.42 \\
Agreeableness & 0.38 \\
\bottomrule
\end{tabular}
\caption{Pearson correlations between Chat and IPIP trait scores.}
\label{tab:correlations}
\end{table}

All correlations were moderate ($r = 0.38$--$0.58$), with Extraversion showing the strongest convergence and Agreeableness the weakest. None reached the threshold typically considered ``strong'' ($r > 0.60$), though all were positive and meaningful.

\begin{figure}[htbp]
    \centering
    \includegraphics[width=\columnwidth]{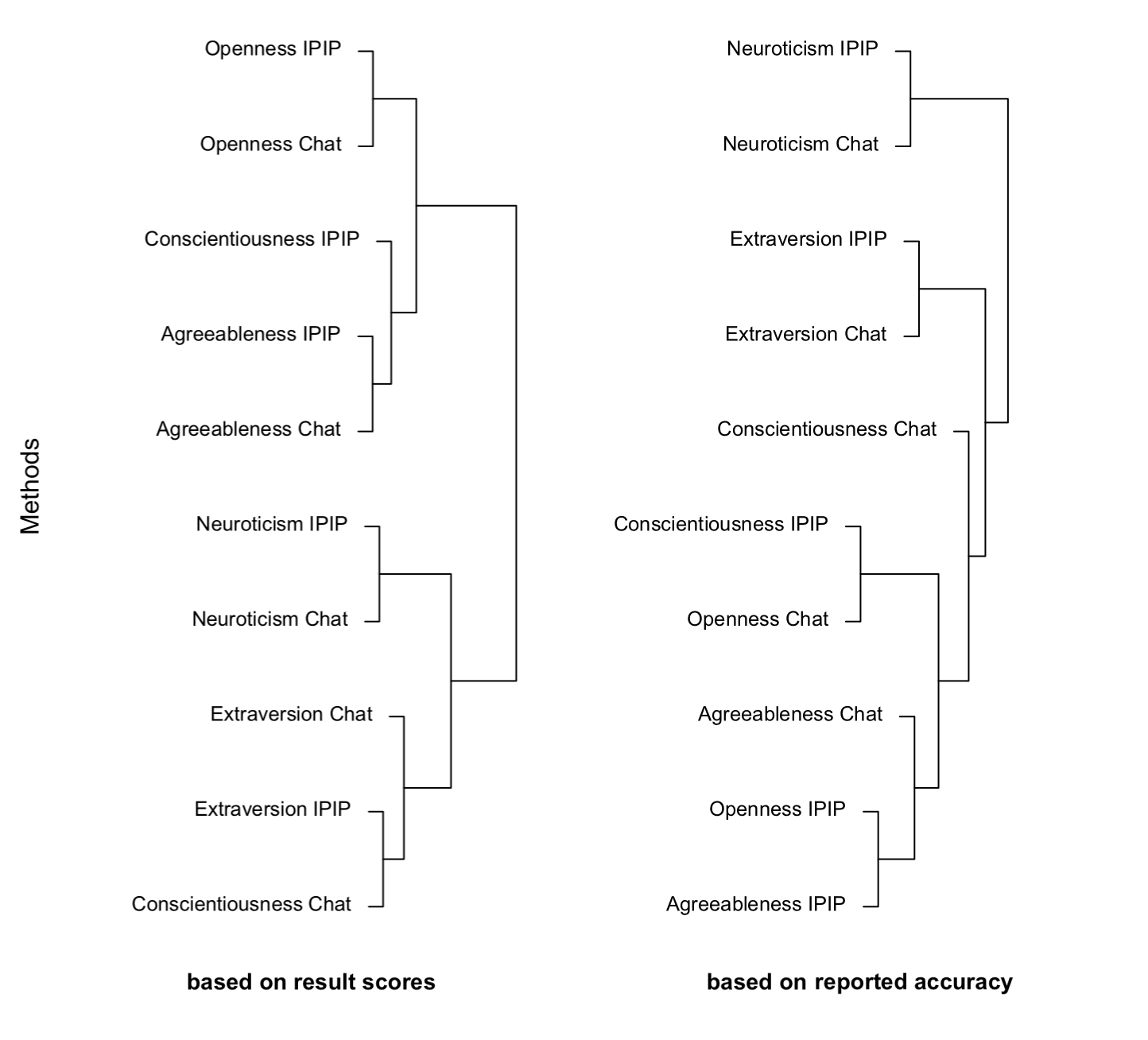}
    \caption{Hierarchical clustering of assessment methods based on result scores (left) and self-reported accuracy ratings (right).}
    \label{fig:methods_clustering}
\end{figure}

Hierarchical clustering of measurement methods (Figure~\ref{fig:methods_clustering}) further reveals the relationship between IPIP and Chat assessments. When methods are clustered by participants' result scores, IPIP and Chat equivalents for most traits cluster together as expected with the exception of Conscientiousness where Chat-derived scores cluster with Extraversion rather than with IPIP Conscientiousness. This suggests method-specific variance in how this trait is captured through conversation versus questionnaire.

When clustered by participants' self-reported accuracy ratings, a different pattern emerges. Neuroticism and Extraversion cluster similarly, while other traits show more dispersed groupings. This suggests that participants experienced comparable confidence in their Neuroticism and Extraversion assessments regardless of method, while accuracy perceptions for other traits were more method-dependent.

\subsection{Perceived Accuracy}

\begin{figure}[htbp]
    \centering
    \includegraphics[width=\columnwidth]{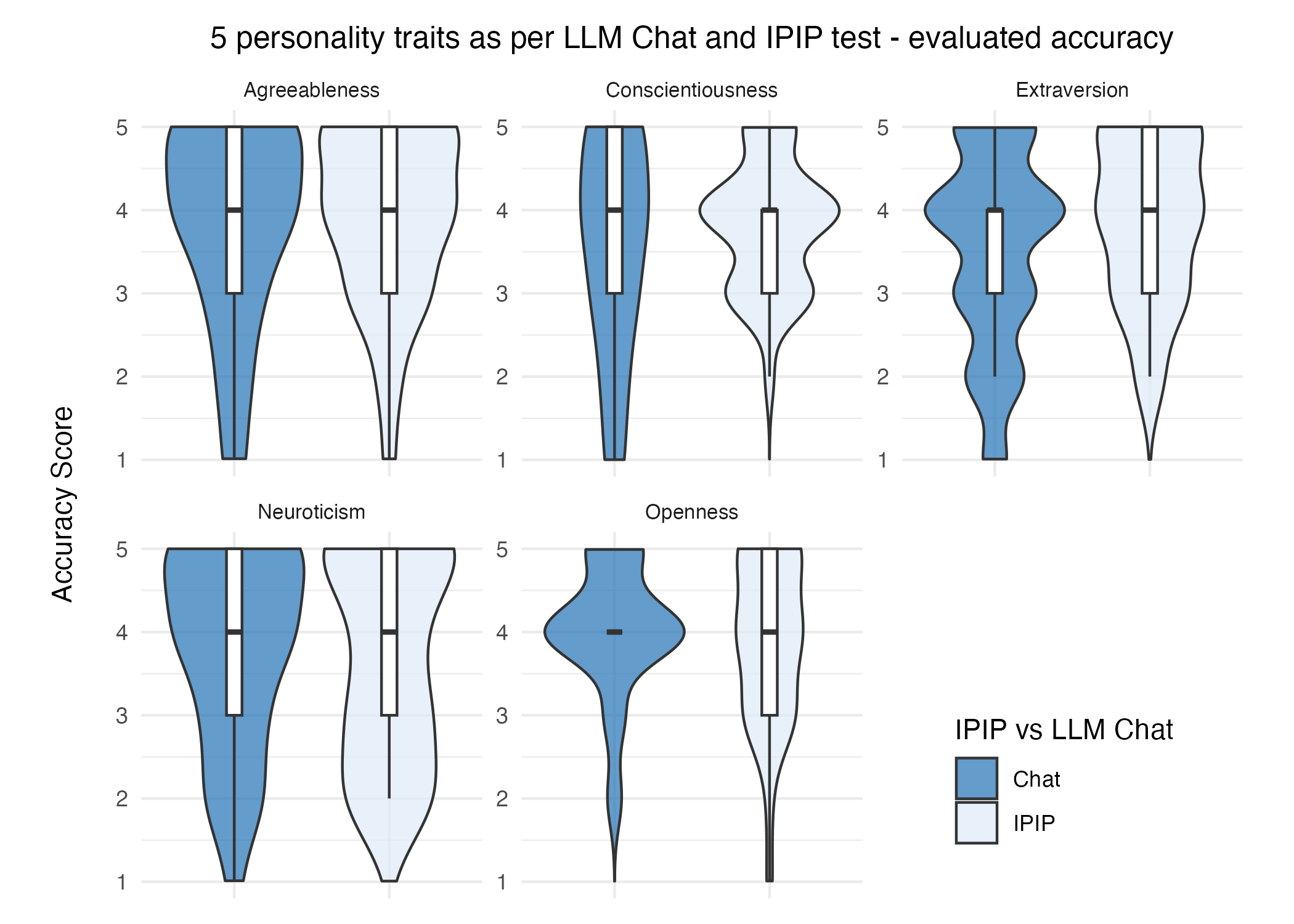}
    \caption{Distribution of participant accuracy ratings by trait and method.}
    \label{fig:accuracy}
\end{figure}

Both methods received high accuracy ratings, with mean scores of 4 out of 5 for each trait across both IPIP and Chat assessments. However, Chat accuracy ratings (Figure~\ref{fig:accuracy}) showed noticeably wider interquartile ranges, indicating greater variability in how participants judged the Chat-generated assessments. This variability was especially pronounced for Conscientiousness and Extraversion.

Importantly, none of the accuracy comparisons reached statistical significance (all $p > 0.05$; see Table~\ref{tab:ttests}), indicating that participants did not systematically perceive one method as more accurate than the other.

Correlations between IPIP accuracy and Chat accuracy ratings were positive for all traits, with Neuroticism and Agreeableness showing moderate positive correlations. This suggests that participants who found IPIP accurate also tended to find Chat accurate.

Overall method preference averaged ``Both were equally accurate,'' with individual responses spanning the full range from ``Chat was much more accurate'' to ``IPIP was much more accurate.''

\subsection{Exploratory Analyses}

PCA conducted on all numeric variables (trait scores, accuracy scores, and method preference) revealed that the first three components explained 62.70\% of total variance (PC1: 32.45\%, PC2: 17.38\%, PC3: 12.88\%). The first five components accounted for 84\% of variance, indicating moderate dimensionality with meaningful variation distributed across multiple constructs.

Analysis of age effects showed no dependency between age group and trait scores. However, older participants were slightly more critical in their accuracy evaluations of both methods.

\begin{figure}[htbp]
    \centering
    \includegraphics[width=\columnwidth]{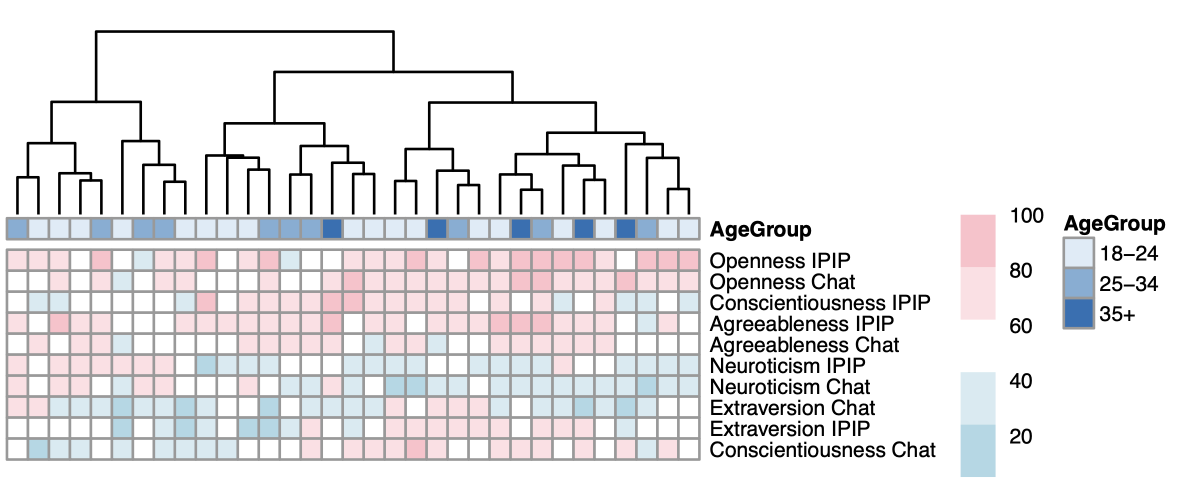}
    \caption{Hierarchical clustering of participants based on Big Five trait scores across both methods.}
    \label{fig:cluster_scores}
\end{figure}

\begin{figure}[htbp]
    \centering
    \includegraphics[width=\columnwidth]{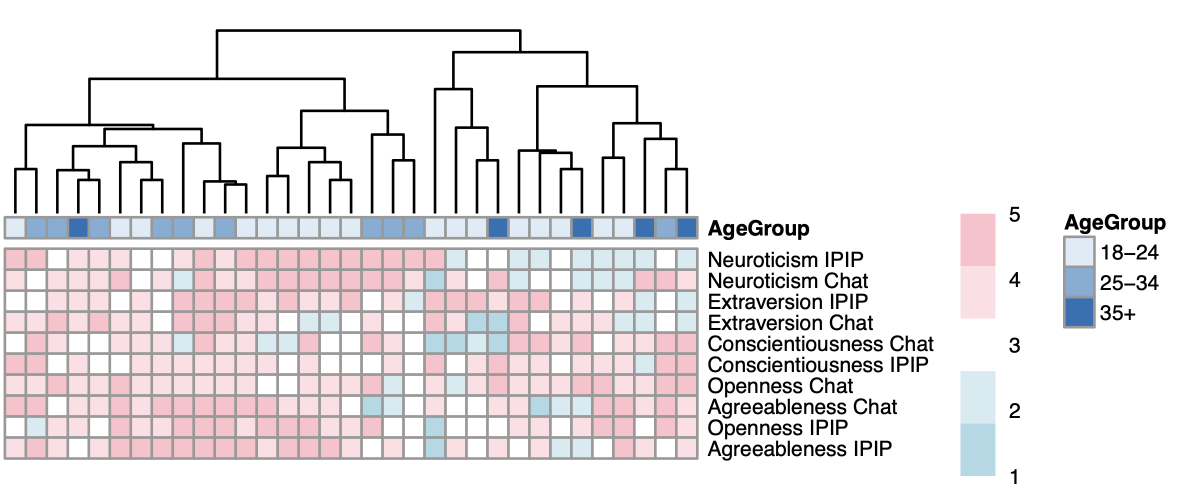}
    \caption{Hierarchical clustering of participants based on self-reported accuracy ratings across both methods.}
    \label{fig:cluster_accuracy}
\end{figure}

To explore individual differences in assessment patterns, we performed hierarchical clustering on participants using both trait scores (Figure~\ref{fig:cluster_scores}) and accuracy ratings (Figure~\ref{fig:cluster_accuracy}).

Clustering based on result scores reveals generally lower values for Neuroticism and Extraversion compared to Openness and Agreeableness across the sample. Score distributions appear visually consistent between IPIP and Chat methods at the participant level, supporting the convergent validity findings. Notably, participant clusters do not align with age groups, suggesting that demographic factors did not systematically influence trait score patterns.

Clustering based on reported accuracy ratings shows greater heterogeneity. While no strong demographic patterns emerge overall, participants who gave lower accuracy ratings for Neuroticism and Extraversion results include a disproportionate number of older participants (3 of 4 in these clusters). However, given our sample size (N=33), this observation remains exploratory and warrants investigation in larger samples.

Correlation analysis between overall method preference and individual trait scores/accuracy ratings revealed minimal relationships, with the exception of a slight negative correlation between final preference and Chat-based Conscientiousness scores.

\section{Discussion}

\subsection{Summary of Findings}

Our results address both research questions:
\begin{enumerate}
    \item LLM-extracted Big Five scores showed moderate correlations with IPIP-50 results ($r = 0.38$--$0.58$). For three of five traits (Conscientiousness, Openness, Neuroticism), the two methods produced statistically equivalent scores. Agreeableness and Extraversion showed significant differences requiring further investigation into method-specific calibration.

    \item Participants perceived both methods as equally accurate, with mean ratings of 4/5 and no significant differences in accuracy judgments across traits. This suggests that conversational assessment is subjectively acceptable to users, even where numerical discrepancies exist.
\end{enumerate}

\subsection{Comparison to Prior Work}

Our correlation coefficients ($r = 0.38$--$0.58$) compare favorably to prior LLM-based personality inference studies:

\begin{itemize}
    \item Peters \& Matz \cite{peters2024}: $r = 0.29$--$0.33$ (zero-shot from Facebook posts)
    \item Piastra \& Catellani \cite{piastra2025}: $r = 0.25$--$0.40$ (ChatGPT-4 from text)
    \item Dai et al. \cite{dai2022}: $r = 0.37$ average (InterviewBERT from interviews)
\end{itemize}

Our higher correlations may reflect the advantages of real-time, guided conversation over passive text analysis. The interactive nature of our approach allows the LLM to probe deeper when responses are ambiguous, potentially capturing personality signals that static text analysis misses.

The discrepancy in Agreeableness scores aligns with challenges noted in prior work. Agreeableness may manifest differently in conversational self-presentation versus self-report questionnaires, where social desirability pressures differ.

\subsection{Use Cases}

These findings support the viability of conversational personality assessment for consumer applications. For dating, hiring, or educational tools, the reasonable convergent validity and high user acceptance suggests that a more natural conversational assessment can complement or even partially replace traditional, static questionnaires.

\subsection{Limitations}

Several limitations constrain interpretation of our findings:

\textbf{Sample size:} With 33 participants, we had adequate power for medium-to-large effects but may have missed smaller true differences. Non-significant results for Conscientiousness ($p = 0.084$) may reflect limited power rather than true equivalence.

\textbf{Population:} Participants were mostly young, educated, and tech-savvy, limiting generalizability to broader populations who may interact differently with conversational AI.

\textbf{Single LLM:} We used one LLM configuration; results may vary across models, prompting strategies, or conversation lengths.

\textbf{Order effects:} All participants completed Chat assessment before IPIP, potentially introducing order effects in accuracy ratings or response patterns.

\textbf{Ground truth:} IPIP-50, while validated, is itself a self-report measure subject to biases. True personality is not directly observable, making validation inherently circular.

\subsection{Future Work}

Future research should address these limitations through larger, more diverse samples and counterbalanced designs. Test-retest reliability of the conversational method should be assessed to determine temporal stability. Comparative studies across different LLM architectures (GPT-5, Claude, Llama) would establish generalizability. Finally, examining which conversational features (length, emotional disclosure, specific topics) predict validity could inform question design optimization.

\section{Conclusion}

This study provides initial validation that LLM-based conversational personality assessment can produce scores largely consistent with traditional IPIP-50 results while achieving comparable user-perceived accuracy. Moderate correlations ($r = 0.38$--$0.58$) and score equivalence for three of five traits suggest that conversational AI can serve as a valid tool for personality assessment in consumer applications. While refinement is needed for Agreeableness and Extraversion measurement, the approach shows promise for more engaging, dynamic alternatives to static questionnaires. Larger-scale replication is recommended to confirm these findings and establish population-level generalizability.

$\,$


\newpage

\onecolumn

\appendix

\section*{Appendix A: System Prompt Architecture}

The conversational assessment system employed two distinct LLM components: (1) a \textbf{Conversational LLM} that conducted 20 guided dialogues with participants, and (2) a \textbf{Scoring LLM} that analyzed all transcripts to generate Big Five trait scores. This appendix details the question design principles, question battery, and prompt engineering approach for both components.

\subsection*{A.1 Question Design Principles}

Questions were designed to elicit personality-relevant behavioral information through natural conversation while avoiding the limitations of traditional self-report measures. Four principles guided question development.

First, questions emphasize behavioral rather than hypothetical framing. Asking ``Tell me about a time when...'' yields richer personality signals than ``What would you do if...'' because participants describe actual behavior rather than idealized self-perceptions. 

Second, questions are specific rather than abstract: ``How do you typically spend your Saturday morning?'' anchors responses in concrete routine, whereas ``How do you like to spend your free time?'' invites vague generalities. 

Third, questions remain low-stakes, avoiding formats that feel like tests or suggest ``right answers.'' Fourth, the battery maintains natural variation by mixing temporal frames (past and present), situational prompts, and reflective questions.

Several constraints further shaped question selection. Questions avoid directly referencing personality traits (e.g., ``Are you organized?''), which tends to elicit socially desirable responses rather than authentic behavioral descriptions. Questions also avoid requiring extended storytelling, which increases participant fatigue, and were checked against IPIP-50 items to minimize overlap with the gold-standard instrument.

Table~\ref{tab:question_transformation} illustrates how these principles transform hypothetical questions into behavioral alternatives.

\begin{table}[H]
\centering
\small
\begin{tabular}{@{}p{6cm}p{8cm}@{}}
\toprule
\textbf{Hypothetical (Avoid)} & \textbf{Behavioral (Preferred)} \\
\midrule
``How do you handle conflict?'' & ``Tell me about the last time you disagreed with someone. What happened?'' \\
\addlinespace
``Do you like philosophical conversations?'' & ``What kind of conversations do you find most engaging?'' \\
\bottomrule
\end{tabular}
\caption{Transformation of hypothetical questions into behavioral alternatives.}
\label{tab:question_transformation}
\end{table}

\subsection*{A.2 Complete Question Battery}

The 20 questions span all five traits (4 questions each). Below is the complete set organized by trait.

\vspace{0.5em}
\noindent\textbf{Agreeableness (Questions 1--4)}
\begin{enumerate}
    \item ``Tell me about the last time someone asked you for help or a favor.''
    \item ``Tell me about the last time you disagreed with someone. What happened?''
    \item ``How do you usually respond when someone is upset or going through a tough time?''
    \item ``Tell me about a time when you had to choose between your own interests and someone else's. What did you do?''
\end{enumerate}

\noindent\textbf{Conscientiousness (Questions 5--8)}
\begin{enumerate}
    \setcounter{enumi}{4}
    \item ``What's one important thing you had to get done this week? How did it go?''
    \item ``How do you keep track of your commitments and responsibilities?''
    \item ``Walk me through how you prepared for something important recently---maybe an exam, presentation, or event.''
    \item ``How do you decide what to work on when you have multiple things competing for your attention?''
\end{enumerate}

\noindent\textbf{Extraversion (Questions 9--12)}
\begin{enumerate}
    \setcounter{enumi}{8}
    \item ``What does your ideal evening look like?''
    \item ``After a long, exhausting day, would you rather spend time alone or call a friend? Why?''
    \item ``When you meet new people at a social event, what's your typical approach?''
    \item ``How do you feel about being the center of attention?''
\end{enumerate}

\noindent\textbf{Neuroticism (Questions 13--16)}
\begin{enumerate}
    \setcounter{enumi}{12}
    \item ``What's something that has annoyed or stressed you out in the past few days?''
    \item ``Tell me about a recent mistake you made. How did you handle it?''
    \item ``When you're facing something uncertain or unpredictable, how do you usually feel?''
    \item ``How do you typically feel at the end of a busy day?''
\end{enumerate}

\noindent\textbf{Openness (Questions 17--20)}
\begin{enumerate}
    \setcounter{enumi}{16}
    \item ``What kind of conversations do you find most engaging?''
    \item ``Where would you like to travel to? Why that place?''
    \item ``Tell me about something new you tried recently. What drew you to it?''
    \item ``When you encounter an idea that challenges your current views, how do you usually react?''
\end{enumerate}

\subsection*{A.3 Question Structure}

Each of the 20 questions have 3 additional text fields as seen in Table~\ref{tab:question_structure} to create a cohesive chat experience.

\begin{table}[H]
\centering
\small
\begin{tabular}{@{}lp{10cm}@{}}
\toprule
\textbf{Field} & \textbf{Description} \\
\midrule
\texttt{trait} & Big Five dimension (e.g., ``Agreeableness'') \\
\texttt{question} & Question text \\
\texttt{guidance} & What to extract, how to probe, tone to maintain, pitfalls to avoid \\
\texttt{criteria} & Primary indicators, follow-up triggers, exit conditions \\
\bottomrule
\end{tabular}
\caption{Question metadata schema for dynamic prompt generation.}
\label{tab:question_structure}
\end{table}

\vspace{0.5em}
\noindent Example Question Entry (Question 2):

\begin{quote}
\small
\ttfamily
trait: "Agreeableness"\\
\\
question: "Tell me about the last time you disagreed with someone. What happened?"\\
\\
guidance:
\begin{adjustwidth}{10pt}{0pt}
"Extract: Approach to conflict (direct/avoidant/compromising), tone during disagreement (calm/heated), concern for relationship vs. being right\\
Probe: If answer is theoretical, ask for specific recent example; if they say they avoid conflict, explore what they do instead\\
Maintain: Non-judgmental about conflict styles\\
Avoid: Don't frame any approach as "better""\\
\end{adjustwidth}
criteria:
\begin{adjustwidth}{10pt}{10pt}
"Primary: Specific disagreement + their approach + concern for other's perspective\\
Follow-up: If only describes outcome, probe their internal experience during conflict\\
Exit: User avoids giving concrete example after two attempts"
\end{adjustwidth}
\end{quote}

\subsection*{A.4 Conversational LLM}

Each conversation is initialized with a dynamically constructed system prompt from the four fields described in A.3. The conversational LLM is instructed to recognize when sufficient information has been gathered.

\vspace{1em}
\noindent\textbf{System Prompt Template:}

\begin{quote}
\small
\ttfamily
You are having a genuine, curious conversation with someone. Your goal is to understand how they think and behave regarding \{trait\} through natural, flowing dialogue.\\
\\
\#\# YOUR OPENING QUESTION\\
"\{question\}"\\
\\
\#\# WHAT YOU'RE CURIOUS ABOUT\\
\{guidance\}\\
\\
\#\# HOW TO HAVE THIS CONVERSATION\\
\\
Think of this as a real conversation with a friend over coffee. The key is to **stay on ONE thread** and know when you have your answer:\\
\\
\#\#\# Step 1: Ask and Listen\\
- Ask your opening question\\
- Really listen to what they share\\
- Recognize if they've already given you what you need\\
\\
\#\#\# Step 2: Go Deeper (Only If Needed)\\
If their answer was vague or surface-level, ask ONE follow-up that explores the same story/example deeper:\\
- "What was that like for you?" (exploring feeling)\\
- "Tell me more about that situation" (getting details)\\
- "How did you feel about that?" (understanding reaction)\\
\\
\#\#\# Step 3: Know When to Stop\\
**CRITICAL - Avoid repetitive questioning:**\\
- If they've given you a concrete example with details → STOP, you have enough\\
- If they've already explained their reasoning or feelings → DON'T ask again in different words\\
- If they've shared a specific story → DON'T ask for another example\\
- If you find yourself wanting to ask essentially the same question → It's time to wrap up\\
\\
**Red flags that you're being repetitive:**\\
- "Can you elaborate more on..." (when they already elaborated)\\
- "What about when..." (introducing a new scenario they didn't mention)\\
- "How do you typically..." (when they already described their typical pattern)\\
- Any question that feels like rephrasing what you just asked\\
\\
**Example of good stopping:**\\
User: "I helped my friend move last weekend. It was exhausting but I felt good about it, like I was really there for them when they needed me."\\
Good: Wrap up now - they've given you behavior, feeling, and reasoning\\
Bad: "And how do you usually feel when people ask for your help?" (repetitive)\\
\\
\{criteria\}\\
\\
\#\# TIMING\\
Brief - about 2-2.5 minutes total. Opening question, then at most ONE follow-up. If they give you a solid answer right away, wrap it up.\\
\\
\#\# WRAPPING UP\\
When you have a clear understanding (usually after 1-2 user responses), thank them naturally and add [CONVERSATION\_COMPLETE].\\
\\
CRITICAL: Only include [CONVERSATION\_COMPLETE] when you're wrapping up WITHOUT asking another question.\\
- Do NOT ask any question in the same message where you include [CONVERSATION\_COMPLETE]\\
- Avoid using the character "?" anywhere in that final response\\
- If user has already responded, follow the guidance to probe deeper only if you truly need clarification\\
- Don't explain the assessment or mention Big Five traits\\
- Stay focused on the current question topic\\
- Exit gracefully if user shows discomfort per exit criteria\\
\\
Begin the conversation now.
\end{quote}

\subsection*{A.5 Scoring LLM}

After all 20 conversations conclude, transcripts are analyzed by a separate Scoring LLM. This separation prevents anchoring effects where mid-conversation score estimates might bias subsequent probing. 

The prompt consists of a brief explanation of the situation, the judgment criteria for Big Five, examples of score moderation, and outlines the expected output. 

It also explicitly instructs to weigh behavioral patterns higher than stated preferences to addresses the limitation that self-descriptions may not align with actual behavior.

\vspace{1em}
\noindent\textbf{Scoring System Prompt:}

\begin{quote}
\small
\ttfamily
You are a psychological assessment expert specializing in Big Five personality profiling. Your task is to analyze conversation transcripts and generate accurate Big Five personality scores based on the IPIP (International Personality Item Pool) framework.\\
\\
\#\# INPUT\\
You will receive a combined text containing 20 conversation transcripts from a single user. These conversations cover various topics designed to reveal personality traits across the Big Five dimensions.\\
\\
\#\# YOUR TASK\\
Analyze the entire conversation text and assign scores for each of the Big Five personality traits on a scale of 0-120, where:\\
- 0-40 = Low (trait is weakly expressed)\\
- 41-80 = Moderate (trait is moderately expressed)\\
- 81-120 = High (trait is strongly expressed)\\
\\
\#\# BIG FIVE TRAITS AND FACETS TO ASSESS\\
\\
\#\#\# 1. OPENNESS TO EXPERIENCE (0-120)\\
Assess intellectual curiosity, aesthetic appreciation, imagination, and willingness to try new experiences.\\
\\
**Key Facets:**\\
- Imagination: Fantasy-prone, daydreaming, creative thinking\\
- Artistic Interests: Appreciation for art, music, poetry\\
- Emotionality: Depth of emotional experience\\
- Adventurousness: Willingness to try new activities, travel preferences\\
- Intellect: Engagement with abstract ideas, philosophical discussions\\
- Liberalism: Readiness to challenge authority and convention\\
\\
**Look for:** Travel preferences (novelty vs. familiarity), intellectual curiosity, engagement with abstract ideas, passion for hobbies, cultural interests, philosophical inclinations.\\
\\
\#\#\# 2. CONSCIENTIOUSNESS (0-120)\\
Assess organization, dependability, work ethic, and goal-directed behavior.\\
\\
**Key Facets:**\\
- Self-Efficacy: Belief in one's ability to accomplish tasks\\
- Orderliness: Preference for organization and structure\\
- Dutifulness: Sense of obligation and responsibility\\
- Achievement-Striving: Ambition and goal-setting\\
- Self-Discipline: Ability to follow through on tasks\\
- Cautiousness: Deliberation before acting\\
\\
**Look for:** Task completion patterns, organizational systems, living space maintenance, planning behaviors, follow-through on commitments, morning routines, project completion rates.\\
\\
\#\#\# 3. EXTRAVERSION (0-120)\\
Assess sociability, assertiveness, energy level, and preference for social interaction.\\
\\
**Key Facets:**\\
- Friendliness: Warmth and approachability with others\\
- Gregariousness: Preference for company vs. solitude\\
- Assertiveness: Social dominance and leadership\\
- Activity Level: Pace and energy of activities\\
- Excitement-Seeking: Need for stimulation\\
- Cheerfulness: Tendency toward positive emotions\\
\\
**Look for:** Ideal evening activities, social recharging preferences, comfort with strangers, group dynamics, social event preferences, energy sources, weekend activities.\\
\\
\#\#\# 4. AGREEABLENESS (0-120)\\
Assess compassion, cooperation, trust, and concern for others.\\
\\
**Key Facets:**\\
- Trust: Belief in others' good intentions\\
- Morality: Straightforwardness and sincerity\\
- Altruism: Concern for others' welfare\\
- Cooperation: Preference for harmony vs. competition\\
- Modesty: Humility vs. self-promotion\\
- Sympathy: Compassion and empathy\\
\\
**Look for:** Relationship values, conflict handling style, helpfulness when asked for favors, trust orientation, friendship qualities valued, responses to criticism, affection expression.\\
\\
\#\#\# 5. NEUROTICISM (0-120)\\
Assess emotional stability, anxiety, stress reactivity, and vulnerability.\\
\\
**Key Facets:**\\
- Anxiety: Tendency to worry and feel tense\\
- Anger: Tendency to experience frustration\\
- Depression: Tendency toward sadness and discouragement\\
- Self-Consciousness: Sensitivity to social evaluation\\
- Immoderation: Difficulty resisting urges\\
- Vulnerability: Susceptibility to stress\\
\\
**Look for:** Recent stressors and reactions, adaptability to plan changes, self-criticism after mistakes, anticipatory anxiety, test/exam stress, compliments that stuck, self-improvement efforts, current life priorities.\\
\\
\#\# SCORING GUIDELINES\\
\\
1. **Read the entire conversation text carefully** before assigning scores\\
2. **Look for patterns** across multiple conversations rather than isolated statements\\
3. **Consider intensity and consistency** of trait expressions\\
4. **Use specific behavioral examples** as evidence (not just self-descriptions)\\
5. **Account for context** - some behaviors may be situational rather than trait-based\\
6. **Be objective** - score what the person demonstrates, not stereotypes or assumptions\\
7. **Balance positive and negative indicators** - absence of high trait expression does not mean low score necessarily\\
8. **Weight behavioral evidence more heavily than self-descriptions**\\
\\
\#\# SCORING ANCHORS\\
\\
**High Scores (81-120):**\\
- Strong, consistent evidence across multiple conversations\\
- Spontaneous mentions of trait-relevant behaviors\\
- Clear behavioral patterns aligned with trait\\
- Examples demonstrate trait even in challenging situations\\
\\
**Moderate Scores (41-80):**\\
- Mixed evidence or moderate expression\\
- Some trait-relevant behaviors but not dominant\\
- Contextual variation in trait expression\\
- Balance of trait-consistent and trait-inconsistent behaviors\\
\\
**Low Scores (0-40):**\\
- Minimal evidence of trait\\
- Counter-trait behaviors dominate\\
- Explicit statements or clear patterns opposing trait\\
- Consistent avoidance of trait-relevant situations\\
\\
\#\# OUTPUT FORMAT\\
\\
Provide your assessment in the following JSON structure. DO NOT include markdown code blocks or any text outside the JSON:\\
\\
\{\\
  "openness": \{\\
    "score": [0-120 integer],\\
    "confidence": "low" | "moderate" | "high",\\
    "key\_evidence": ["Brief behavioral indicator 1", "Brief behavioral indicator 2", "Brief behavioral indicator 3"],\\
    "reasoning": "2-3 sentence explanation of score based on patterns observed"\\
  \},\\
  "conscientiousness": \{\\
    "score": [0-120 integer],\\
    "confidence": "low" | "moderate" | "high",\\
    "key\_evidence": ["Brief behavioral indicator 1", "Brief behavioral indicator 2", "Brief behavioral indicator 3"],\\
    "reasoning": "2-3 sentence explanation of score based on patterns observed"\\
  \},\\
  "extraversion": \{\\
    "score": [0-120 integer],\\
    "confidence": "low" | "moderate" | "high",\\
    "key\_evidence": ["Brief behavioral indicator 1", "Brief behavioral indicator 2", "Brief behavioral indicator 3"],\\
    "reasoning": "2-3 sentence explanation of score based on patterns observed"\\
  \},\\
  "agreeableness": \{\\
    "score": [0-120 integer],\\
    "confidence": "low" | "moderate" | "high",\\
    "key\_evidence": ["Brief behavioral indicator 1", "Brief behavioral indicator 2", "Brief behavioral indicator 3"],\\
    "reasoning": "2-3 sentence explanation of score based on patterns observed"\\
  \},\\
  "neuroticism": \{\\
    "score": [0-120 integer],\\
    "confidence": "low" | "moderate" | "high",\\
    "key\_evidence": ["Brief behavioral indicator 1", "Brief behavioral indicator 2", "Brief behavioral indicator 3"],\\
    "reasoning": "2-3 sentence explanation of score based on patterns observed"\\
  \},\\
  "overall\_assessment": "2-3 sentence summary of the person's overall personality profile and key characteristics"\\
\}\\
\\
\#\# IMPORTANT NOTES\\
\\
- **DO NOT** simply average responses - weight behavioral evidence more heavily than self-descriptions\\
- **DO NOT** be overly influenced by socially desirable responses\\
- **DO** consider that people may have limited self-awareness in some areas\\
- **DO** look for implicit indicators (what they show) vs explicit claims (what they say)\\
- **BE CONSISTENT** in your scoring approach across all five traits\\
- If information is insufficient for a particular trait, indicate "low" confidence and score conservatively toward the middle range (50-70)\\
- **CRITICAL**: Your response must be ONLY valid JSON with no additional text, markdown formatting, or code blocks\\
\\
Now, please analyze the following conversation transcripts and provide Big Five personality scores:
\end{quote}

\section*{Appendix B: Ethics Statement}

This research was conducted with strict adherence to ethical principles, prioritizing the privacy of human participants. All participants were at least 18 years of age and provided informed consent prior to data collection. The consent process disclosed the study's purpose (validating AI-based personality profiling), described all tasks (20 AI conversations, IPIP-50 questionnaire, accuracy ratings), and enumerated the data collected (conversation transcripts, questionnaire responses, demographic information).

Participation was voluntary, and participants were informed of their right to withdraw at any time without penalty by contacting the research team. Confidentiality was ensured by restricting data access to the research team and university supervisor, with no personally identifiable information shared with third parties at any stage of the experiment. All personal identifiers are to be deleted by January 31, 2026. Fully anonymized data may be retained for further research purposes.

Data handling procedures align with GDPR requirements and University of Tartu ethical guidelines. Published results report only anonymized, aggregate findings. The complete informed consent document is available from the corresponding author upon request.

\end{document}